\documentclass[letterpaper, 10 pt, conference]{ieeeconf}  

\IEEEoverridecommandlockouts                              

\usepackage{amsmath} 
\usepackage{xcolor}
\usepackage{graphicx}
\usepackage{subcaption}
\usepackage{siunitx}
\usepackage{xspace}

\usepackage{hyperref}  

\title{\LARGE \bf
Generative adversarial imitation learning for robot swarms:\\
Learning from human demonstrations and trained policies
}

\author{Mattes Kraus$^{1}$ and Jonas Kuckling$^{1,2,3}$
\thanks{$^{1}$ University of Konstanz, Department of Computer and Information Science, Konstanz, Germany}%
\thanks{$^{2}$ Cluster of Excellence ``Centre for the Advanced Study of Collective Behavior'', University of Konstanz}%
\thanks{$^{3}$ Zukunftskolleg, University of Konstanz}
\thanks{*JK acknowledges funding from the Carl-Zeiss-Stiftung and from the Zukunftskolleg. The research was supported by the Cluster of Excellence ``Centre for the Advanced Study of Collective Behavior'', under Germany's Excellence Strategy – EXC 2117 – 422037984.}
\thanks{The experiments were designed by MK and JK and performed by MK. Real robot experiments were performed by MK and JK. The paper was written and edited by MK and JK, all authors read and approved the final version. The research was directed by JK.}
\thanks{\tt\small jonas.kuckling@uni-konstanz.de}}

\newcommand{\change}[1]{#1}

\newcommand{\MStandingStill}{\textsc{Standing Still}\xspace}
\newcommand{\MFullSpeed}{\textsc{Full Speed}\xspace}
\newcommand{\MControlledSpeed}{\textsc{Controlled Speed}\xspace}
\newcommand{\MAggregation}{\textsc{Aggregation}\xspace}
\newcommand{\MDispersion}{\textsc{Dispersion}\xspace}
\newcommand{\MForaging}{\textsc{Foraging}\xspace}

\begin{document}

\maketitle
\thispagestyle{empty}
\pagestyle{empty}

\begin{abstract}
In imitation learning, robots are supposed to learn from demonstrations of the desired behavior.
Most of the work in imitation learning for swarm robotics provides the demonstrations as rollouts of an existing policy.
In this work, we provide a framework based on generative adversarial imitation learning that aims to learn collective behaviors from human demonstrations.
Our framework is evaluated across six different missions, learning both from manual demonstrations and demonstrations derived from a PPO-trained policy.
Results show that the imitation learning process is able to learn qualitatively meaningful behaviors that perform similarly well as the provided demonstrations.
Additionally, we deploy the learned policies on a swarm of TurtleBot~4 robots in real-robot experiments.
The exhibited behaviors preserved their visually recognizable character and their performance is comparable to the one achieved in simulation.
\end{abstract}

\section{Introduction}

A robot swarm is a decentralized, homogeneous multi-robot systems, in which each robots acts autonomously based only on the information that is locally available to it~\cite{BraFerBirDor2013SI,DorBirBra2014SCHOLAR}.
These properties lend themselves to the creation of multi-robot systems that exhibit some desirable properties, such as scalability, flexibility and robustness~\cite{DorBirBra2014SCHOLAR}.
While robot swarms have shown to be able to tackle a wide range of  missions~\cite{BraFerBirDor2013SI,SchUmlSenElm2020FRAI,DorTheTri2021PIEEE}, designing control software for any particular mission remains a key challenge in swarm robotics~\cite{FraBir2016FRAI,Kuc2023FRAI}.
This is due to the fact that the swarm needs to be programmed on the individual level, but the desired collective behavior is emergent from the numerous and often unpredictable interactions of the robots.
Manual design approaches therefore often rely on trial-and-error, and therefore depend on the skill of the designer~\cite{BirLigBoz-etal2019FRAI}.

Automatic methods transform the design problem into an optimization problem~\cite{BirLigBoz-etal2019FRAI}.
Instead of finding suitable control software to address a specific mission, an optimization algorithm searches for the instance of control software that maximizes some (mission-specific) performance measure.
It is generally assumed that the instance of control software that maximizes the mission-specific performance measure also solves the mission in a satisfactory way.
However, designing these performance measures is itself a challenging problem.
If they only measure the desired outcomes, they might not provide enough information during early stages of the design process~\cite{LehSta2011EC,DonBreMouEib2015FRAI,DivHam2015gecco}.
Contrarily, if the designer also encodes ``how'' the behavior should be achieved, 
the performance measure becomes prone to reward hacking~\cite{NgHarRus1999icml}.
As a result, high-scoring behaviors might not correspond to desired ones.

Imitation learning approaches offer an alternative way to specify desired behaviors.
Instead of providing an explicit performance measure, the designer provides \emph{demonstrations} of the desired behavior.
The imitation learning algorithm will then derive a policy that exhibits the same distribution of \emph{features} as the demonstrations.

In this work, we propose an imitation learning framework for robot swarms that uses generative adversarial imitation learning (GAIL)~\cite{HoErm2016neurips}.
We focus on swarm-level features to capture emergent collective dynamics, rather than focusing on the behavior of individual agents.
Demonstrations are provided by human operators that can control the robot swarm through a custom-developed demonstration tool.
To evaluate the impact of human-provided demonstrations, we also consider imitation learning from demonstrations that are derived from policies trained with proximal policy optimization (PPO)~\cite{SchWol-etal2017arxiv}.
The framework is further validated through real-robot experiments, deploying the learned policies on a real swarm of TurtleBot~4 robots.

\section{Related Work}
Works that consider imitation learning in swarm robotics usually employ forms of behavior cloning, explicit feature matching, inverse reinforcement learning, or generative adversarial imitation learning.
%
In behavior cloning, the demonstrations are used to collect sets of state-action (or observation-action) pairs.
The problem is then cast as a supervised learning problem, i.e., the policy is trained to predict the corresponding actions to a states given as input to it~\cite{MatBro2025hitlrl,RahWhi-etal2023acc,ZhoPhi-etal2019iros}.
%
In the case of explicit feature matching, the demonstrations are used to compute a set of features (on the agent- or swarm-level) describing the desired behavior.
The imitation learning process than aims to explicitly minimize the difference between the features in the demonstrations and the features of the generated behaviors~\cite{AlhAbdHau2022ants,AlhAbdHau2025evogp}.
%
In inverse reinforcement learning approaches, the difference penalty for feature distributions is not explicitly given.
Instead, the features are used to learn a reward function, which might assign different importance to some features~\cite{SosKhuZouKoe2017aamas,GhaKucGarBir2023icra,SzpGarBir2024iros}.

In generative adversarial imitation learning approaches, no explicit objective function exists.
Instead, a discriminator and generator (the policy) are trained in parallel~\cite{HoErm2016neurips}.
The policy generates state-action pairs and the discriminator tries to decide if any given state-action pair was part of the demonstration or is generated.
The discriminator is trained to decrease the classification error and the policy is rewarded if it generates state-action pairs that the discriminator incorrectly classifies as originating from the demonstration.
A first work of generative adversarial imitation learning in multi-robot systems is Turing learning~\cite{li_turing_2016}.
Li et al. use co-evolution to evolve one discriminator network and generator model per imitating agent.
The discriminator only has access to the current linear and angular speed of the agent.
Wu et al. use state-action pairs to train the discriminator using OpenAI-ES, which uses state-action pairs of each agent~\cite{wu_adversarial_2025}.
As policy, they employ modular, deep attention networks.
Agunloye et al. use multi-agent generative adversarial imitation learning to learn swarm behaviors for a swarm of UAVs~\cite{AguRamSoo2024iros}.
In order to generate homogeneous policies, the authors employed MA-PPO with parameter sharing.
The collective behavior is modeled through agent-level features that compute the distances to all relevant elements in the environment.

One main challenge in imitation learning for swarm robotics remains the way demonstrations are provided.
Most of the above mentioned works assume that a policy already exists that generates the desired behavior.
In practical terms, this leads to a bootstrapping paradox: If the policy already exists, there is no need to imitate it and if it cannot be generated easily, the imitation learning process has no access to expert demonstrations.
Only few works address directly the problem of a human operator providing demonstrations.
Notably, Gharbi et al.~\cite{GhaKucGarBir2023icra} and Szpirer et al.~\cite{SzpGarBir2024iros} use demonstrations that include only the final position of all robots in the environment.
These demonstrations can easily be provided by a human operator prior to the imitation learning process.
Mattson and Brown tackled the problem of giving demonstrations to a whole swarm by designing a round-robin system~\cite{MatBro2025hitlrl}.
The human operator would control a single robot, while the remaining robots in the swarm were controlled by random policies.
The human-controlled robot would then use behavior cloning to derive a policy mimicking the demonstration.
In the following steps, the human would switch through the other robots, providing demonstrations for each of them while the remaining robots use their learned policies.

\section{Methodology}
In order to apply imitation learning techniques to swarm robotics, a few considerations need to be made.
Features can be defined either on the agent-level or on a collective level.
While agent-level features are more closely connected to the policy, they cannot capture all properties of the swarm.
Therefore, we focus on swarm-level features (see Sec.~\ref{sec:features}).
This allows us to model the problem as a single-agent imitation learning problem (see Sec.~\ref{sec:swarm-gail}).

\subsection{Robot model}
\begin{table}[tb]
    \centering
    \caption{The simplified model for the TurtleBot~4 robot.}
    \label{tab:robot-model}
    \begin{tabular}{c|c|c}
        Sensor & Range & Dim. \\  \hline 
        Speed $|v|$ & $[0, 0.31m]$ \unit{\meter/\second} & 1 \\
        LiDAR & $[0, 200]$ \unit{\centi\meter} & 5 \\
        Color & $\{0;1\}$ & 3 \\
        Bumper & $\{0;1\}$ & 1 \vspace{1em} \\ 
        Actuator & Range & Dim. \\ \hline
        Linear vel. $v$ & $[0, 0.31]$ \unit{\meter/\second} & 1 \\
        Angular vel. $\omega$ & $[-1.9,1.9]$ \unit{\radian/\second} & 1 \\
    \end{tabular}
\end{table}

We provide demonstrations and design policies for a swarm of three TurtleBot~4 robots\footnote{\url{https://clearpathrobotics.com/turtlebot-4/}}.
These robots are commonly used in robotics research and provide several sensors, including infra-red proximity sensors, a 2-d LiDAR, and a bumper for collision detection.
In order to reduce the dimensionality of the sensor data, we consider a simplified robot model (see Table~\ref{tab:robot-model}).
The robot is modeled as a differential drive agent with linear velocity $v$ and angular velocity $\omega$.
%
For sensing, the robot has access to its own current absolute speed $|v|$. 
It can also sense neighboring robots through its LiDAR.
The LiDAR readings are aggregated into five equal sectors (front-right, right, front-left, left, back) that cover \qty{72}{\degree} each.
For each sector, only the closest robot is considered.
If a sector does not contain any other robot, the reading is capped at \qty{200}{\centi\meter}.
In order to be able to react to different zones in the environment, we model that the robots are equipped with ground sensors.
The ground sensors can detect three distinct gray-scale colors: black, white and gray.
The ground color is provided as a one-hot encoded, three-dimensional vector.
Lastly, robots are able to access their bumper to detect collisions with other robots or the worlds.

\subsection{Features for collective behavior}
\label{sec:features}

We select five features to describe the collective behavior: average speed, grouping, coverage, color travel time, and color visit frequency.
These features are computed at every time step during the execution of the mission.
\emph{Average speed} is defined as $\frac{1}{N}\sum_{i=1}^N |v_i|$, with $N$ as the swarm size and $v_i$ as the linear velocity of robot $i$.
The \emph{grouping} feature describes the average distance of the robots to the swarm center.
It is computed by $\frac{1}{N}\sum_{i=1}^N |x_i - \overline{x}|$ with $N$ as the swarm size, $x_i$ as the current position of robot i and $\overline{x}$ as the center of mass position of the swarm.
In order to compute \emph{coverage}, the arena is divided into a $4\times4$ grid.
Each tile counts the time since it was last visited by a robot.
The grouping and coverage metric are taken from~\cite{jeston-fenton_visualisation_2022}.
We also include two metrics to describe the behavior of the robots with respect to the colored zones in the environment.
For the \emph{color visit frequency}, we keep track of the sequence of ground colors that each robot visits.
As gray is the default color of the environment, we focus only on white and black floor patches.
Every time a robot enters a colored patch, we increment a counter based on the current color and the previous visited ground color.
This results in four counters, one for each combination of white-white, white-black, black-white, and black-black.
The \emph{color travel time} is defined as the average time it takes a robot to change between floor patches of any color.
This is one feature describing the average time of all of the four possible combinations.
This results in a total of \num{23} swarm-level features for every time step.

\subsection{Demonstration Tool/Simulator}
\begin{figure}[tb]
  \centering
  \begin{subfigure}{\linewidth}
    \includegraphics[width=\linewidth]{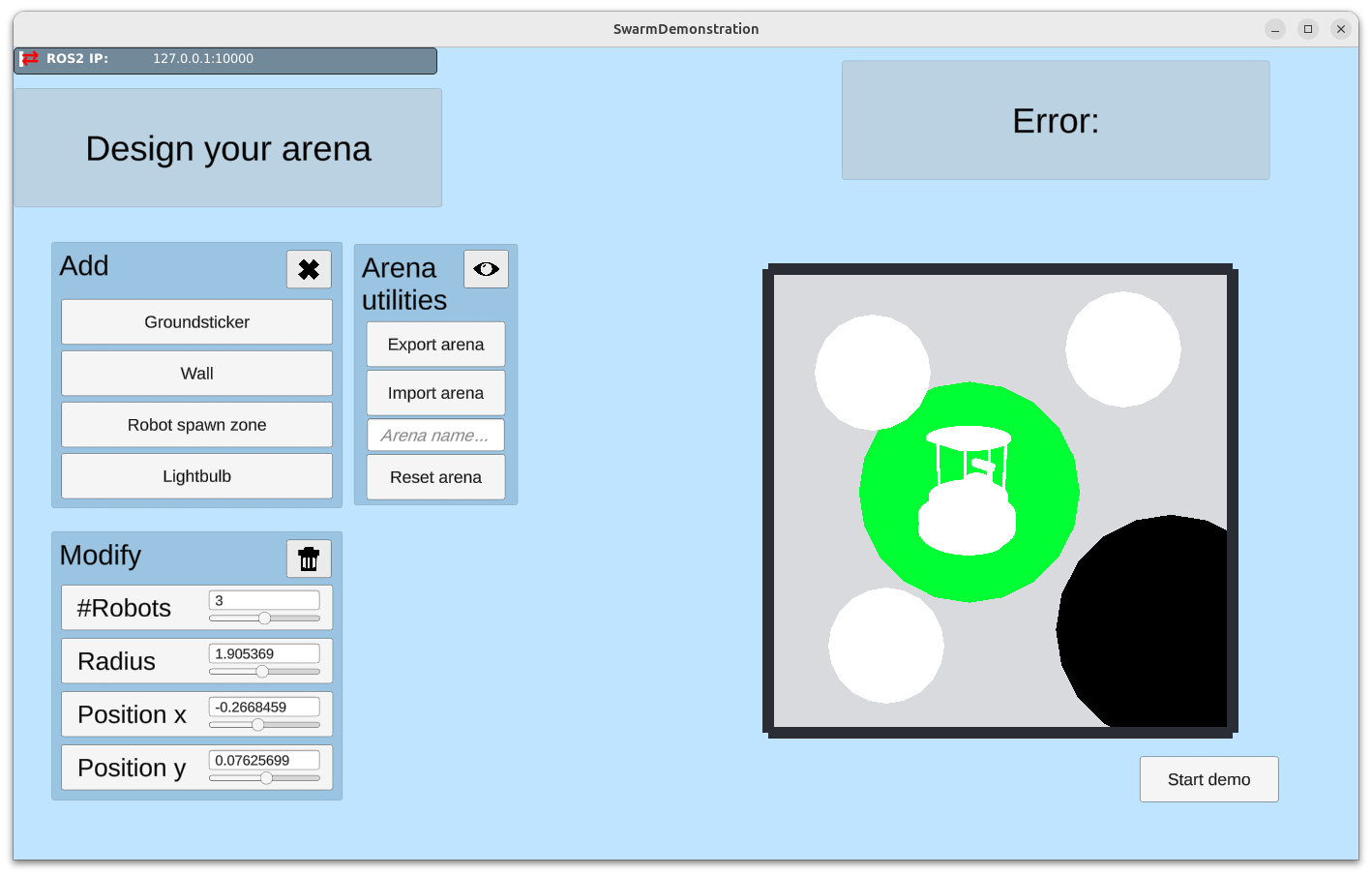}
    \caption{Arena building stage}
    \label{fig:arena-builder}
  \end{subfigure}
  \begin{subfigure}{\linewidth}
      \includegraphics[width=\linewidth]{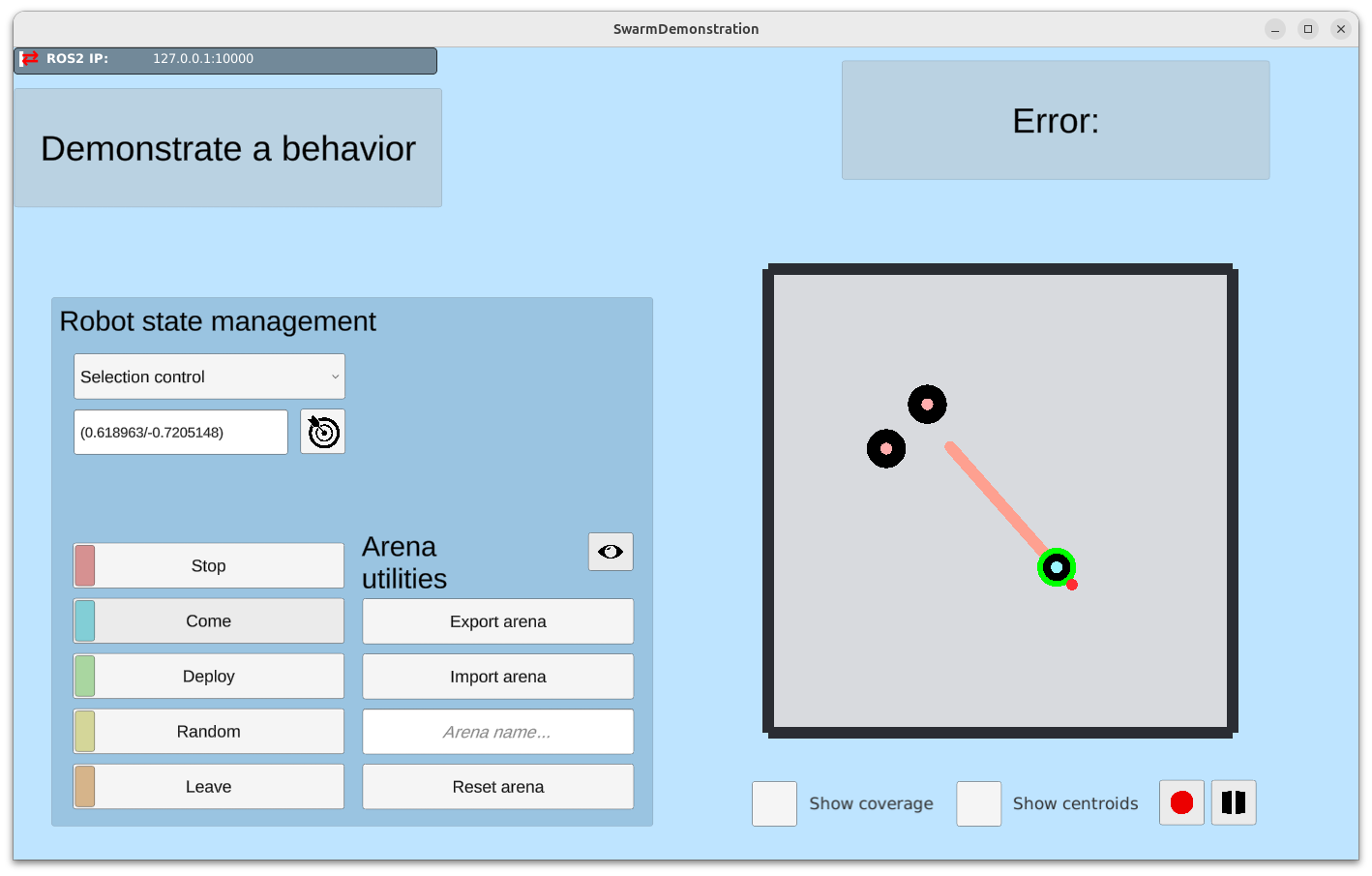}
      \caption{Demonstration building stage}
      \label{fig:demonstration-builder}
  \end{subfigure}
  \caption{Two screen captures from our demonstration tool. The user can build an experimental arena and provide demonstrations of swarm behaviors.}
  \label{fig:demonstration-tool}
\end{figure}
   
In order to provide demonstrations, we develop a demonstration tool (see Fig.~\ref{fig:demonstration-tool}).
The tool is based on the Unity game engine\footnote{\url{https://unity.com/products/unity-engine}} and is available for download\footnote{\url{https://github.com/CPS-Konstanz/SwarmGAIL}}.
The demonstration tool allows to design different environments for the swarm (see Fig.~\ref{fig:arena-builder}).
The user can modify the $\qty{4}{\meter}\times\qty{4}{\meter}$ arena by placing walls and black or white ground patches.
In addition, the user can specify spawn zones for the robots.
In the beginning of the experiment, the robots will randomly placed within these zones.
These configured arenas can be exported and imported so experiments can be easily repeated.

After designing the environment for the swarm, the user can provide demonstrations (see Fig.~\ref{fig:demonstration-builder}).
Using the demonstration tool, the user can give high-level commands to the swarm or individual robots in it.
\change{Prior research has shown that high-level interaction commands result in better demonstrations than when users are presented with low-level control~\cite{HusNguAbb2025PTRSA}}.
The available interaction commands are adapted from Kolling et al.~\cite{kolling_towards_2012}.
Kolling et al. define two control modes: \emph{selection control} and \emph{beacon control}.
In selection control, the user can select which robots shall change their behavior by clicking on them.
Beacon control is an environmental control method, where the user does not directly change the behavior of the robots.
Rather, they can place beacons, trigger zones that are connected a specific behavior.
Whenever a robot enters the beacon, it immediately switches to the connected behavior of the beacon.

Kolling et al. defined seven behaviors that each robot could execute: \textit{Stop}, \textit{Random}, \textit{Come}, \textit{Leave}, \textit{Heading}, \textit{Rendezvous}, and \textit{Deploy}.
However, not every behavior was used equally often by the operators.
Therefore, we did not implement \textit{Rendezvous} and \textit{Heading}.
The remaining behaviors were implemented as follows:
When executing the \textit{Stop} behavior, the robot is standing still.
In the \textit{Random} behavior, the robot performs a random walk in the form of ballistic motion.
The robot moves straight with maximal speed until it encounters an obstacle.
It then turns in place for a random number of time steps before continuing to move straight.
For the \textit{Come} and \textit{Leave} behaviors, a target location within the arena needs to be chosen.
In the \textit{Come} behavior, the robot drives with maximal speed towards the target.
When approaching the target, it progressively slows down.
When the target is within a radius of \qty{20}{\centi\meter}, the robot stops.
The \textit{Leave} behavior repels the robot from the target.
On collision it switches its behavior to \textit{Random}.
In the \emph{Deploy} behavior, the robot computes a Voronoi diagram with each robot position as site.
The robot drives with maximal speed to the \emph{centroid} of its assigned cell.
On collision with other robots, it changes its direction slightly to the right, to avoid further collisions.

In order to collect the user demonstrations, the demonstration tool is able to provide a light-weight simulation of the robot behavior.
In order to minimize the difference in simulation contexts between demonstration and design, we also use the demonstration tool as simulator for the imitation learning process.
Communication between the GAIL library and the demonstration tool is facilitated using ROS~2 and the Unity Robotics Hub\footnote{\url{https://github.com/Unity-Technologies/Unity-Robotics-Hub}}.

\subsection{SwarmGAIL}
\label{sec:swarm-gail}

\begin{figure}[tb]
  \centering
  \includegraphics[width=0.4\textwidth]{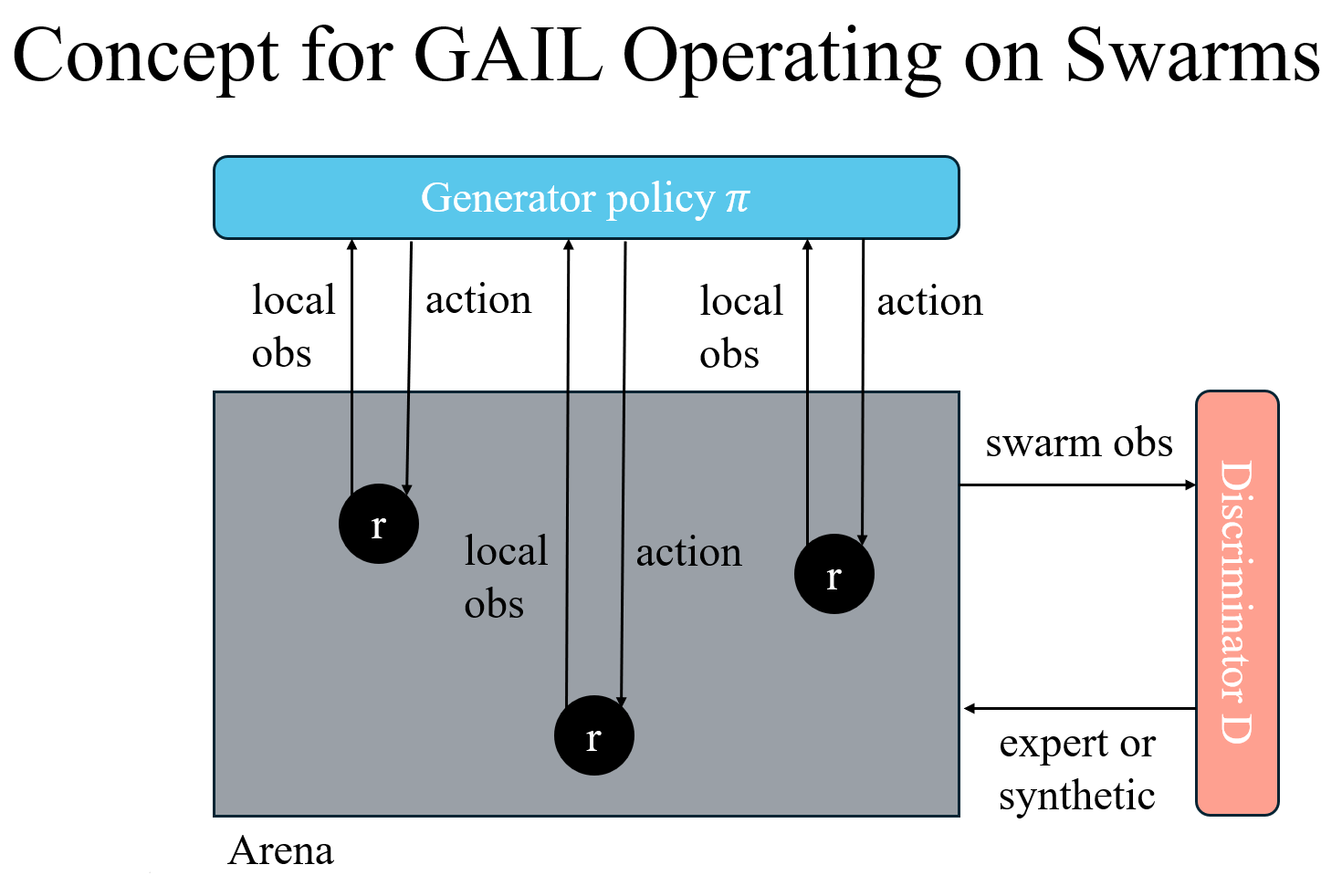}
  \caption{Information flow in our SwarmGAIL implementation. The discriminator receives swarm-level features as observation. The policy is used in a round-robin fashion to provide actions to all robots based on their local observations.
  }
  \label{fig:gail}
\end{figure}

We reduce the robot swarm imitation learning problem to a single-agent generative adversarial imitation learning problem.
For our imitation learning algorithm, we use a slightly adjusted version of GAIL~\cite{HoErm2016neurips} (see Fig.~\ref{fig:gail}).
The algorithm learns a single policy (generator) that controls each robot in the swarm in a round-robin fashion.
The policy takes as input the observations generated by the robot model (see Table~\ref{tab:robot-model}) and outputs the actuator commands for the focal robot.
As the policy is decentralized and only depends on the observations of a single robot, we can later deploy a copy of the policy on each robot.


The discriminator aims to distinguish between demonstrated and generated behavior.
As input, we provide feature-action pairs (with the swarm-level features defined in Section~\ref{sec:features}).
Notably, the features do not include the local observations of the current focal robot.
This ensures that the discriminator only tries to make a classification based on the swarm behavior and not inadvertent differences in the observations between demonstrations and generated behaviors.



We implemented this adjusted GAIL version in Python~3 using the \texttt{imitation} library~\cite{GleTau-etal2022arxiv}.
The discriminator is a multi-layer perceptron with two hidden layers of size \num{32} each.
The policy is a multi-layer perceptron with two hidden layers of size \num{64} each and is trained using PPO~\cite{SchWol-etal2017arxiv}.
For technical reasons, the policy and discriminator expect the same observation space.
We therefore consider a joint observation vector that combines local observations and swarm features.
We overwrite the unnecessary parts of the observation vector with \num{0} before supplying the vector to either the discriminator or the policy.

\section{Experimental setup}

\subsection{Missions}

\begin{figure}[tb]
    \centering
    \begin{subfigure}[t]{.3\linewidth}
        \includegraphics[width=\textwidth]{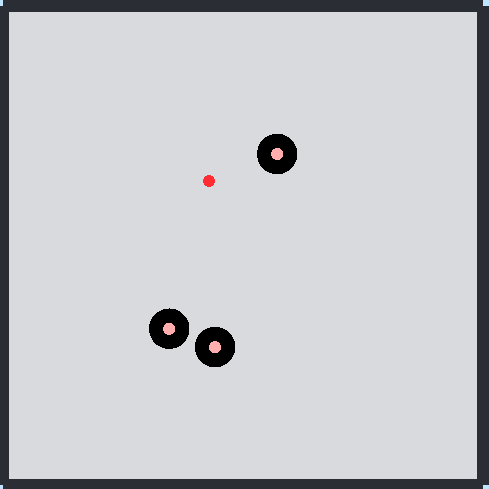}
        \caption{Standing still}
        \label{fig:standing-still}
    \end{subfigure}
    \begin{subfigure}[t]{.3\linewidth}
        \includegraphics[width=\textwidth]{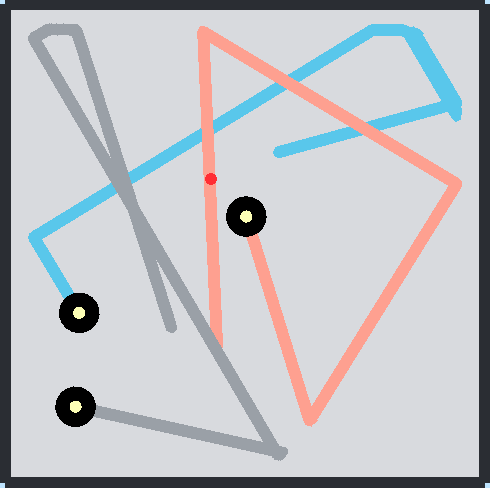}
        \caption{Full speed and controlled speed.}
        \label{fig:full-speed}
    \end{subfigure}
    \begin{subfigure}[t]{.3\linewidth}
        \centering
      \includegraphics[width=\textwidth]{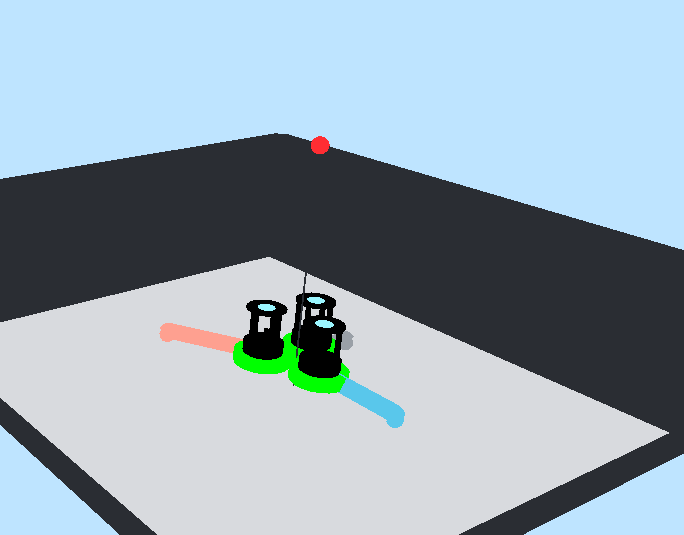}
      \caption{Side view.}
      \label{fig:aggregation-3d}
    \end{subfigure}
    \begin{subfigure}[t]{.3\linewidth}
        \includegraphics[width=\textwidth]{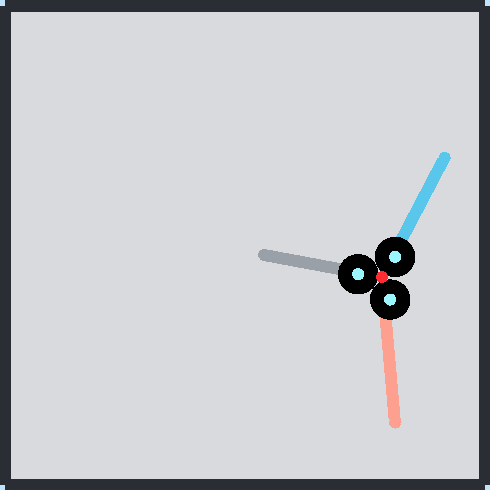}
        \caption{Aggregation}
        \label{fig:aggregation}
    \end{subfigure}
    \begin{subfigure}[t]{.3\linewidth}
        \centering
      \includegraphics[width=\textwidth]{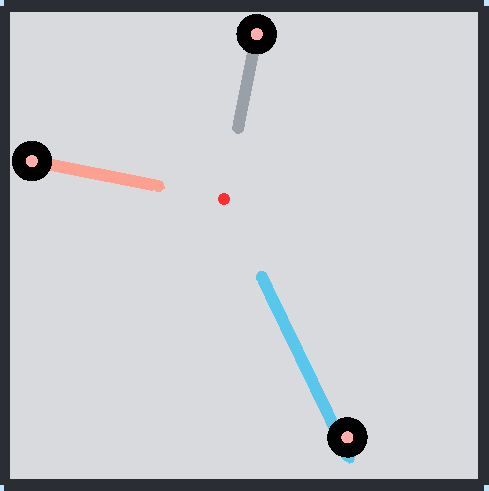}
      \caption{Dispersion}
      \label{fig:dispersion}
    \end{subfigure}
    \begin{subfigure}[t]{.3\linewidth}
        \centering
      \includegraphics[width=\textwidth]{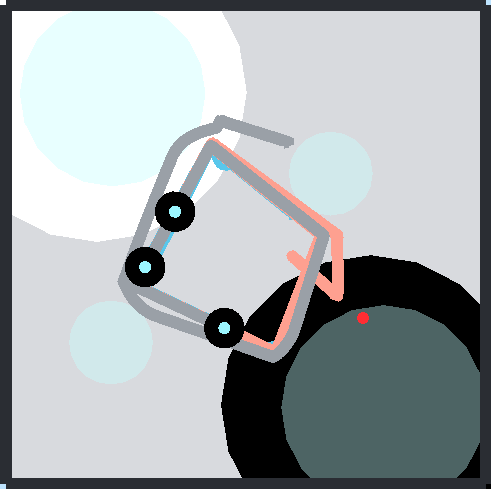}
      \caption{Foraging}
      \label{fig:foraging}
    \end{subfigure}
    \caption{Representations of the demonstrated behavior for all considered missions. In the case of \MFullSpeed and \MControlledSpeed, the trajectories taken by the robots look visually similar, but the velocity of the robots differs.}
    \label{fig:missions}
\end{figure}

To test our imitation learning framework, we consider six missions: \MStandingStill, \MFullSpeed, \MControlledSpeed, \MAggregation, \MDispersion and \MForaging (see Fig.~\ref{fig:missions}).
For each mission, we define a general strategy to provide demonstrations and, additionally, a reward function that we assume is maximized by our demonstrations.

The missions \MStandingStill, \MFullSpeed, \MControlledSpeed, \MAggregation and \MDispersion take place in an empty arena.
The robots spawn approximately in the middle of the environment.
In \MStandingStill, the robots should not move forward (stationary rotation is allowed).
The demonstration is performed by selecting all robots and assigning the \textit{Stop} behavior.
The reward function is defined as $R_s(t)=-\bar{v}(t)$, where $\bar{v}(t)$ is the average linear velocity of the swarm at time step $t$.
In \MFullSpeed, robots should move as fast as possible.
Therefore, the reward function for this mission is defined as $R_f(t)=\bar{v}(t)$.
For the demonstrations, all robots are selected and we apply the \textit{Random} behavior to them.
For \MControlledSpeed, the robots should move at a constant velocity $v_\mathit{target}=\qty{0.1}{\meter/\second}$.
As none of the behaviors in our demonstration tool allows to explicitly set the linear velocity, we reduced the maximum allowed velocity for the demonstrations.
Then we selected all robots and applied the \textit{Random} behavior to them.
The reward function for this mission is defined as $R_c(t)=|v_\mathit{target}-\bar{v}(t)|$.

In \MAggregation, the robots should aggregate in the arena, but no particular spot is required for the aggregation.
The reward function is defined as $R_a(t)=1/\bar{g}$, where $\bar{g}$ is the grouping feature, i.e., the average distance of the robots to the center of mass of the swarm.
Demonstrations were provided by selecting all robots and assigning them the \textit{Come} behavior with the target being chosen as approximately the center of mass of the swarm. 
Similarly, in \MDispersion, the robots should spread out as far as possible in the arena.
The reward function is defined as $R_d(t)=-1/\bar{g}$.
Demonstrations are provided by selecting all robots and assigning them the \textit{Leave} behavior.
\MForaging is the most complex behavior that we considered.
The environment contains a white area (the nest) and a black area (the source) and the robots spawn in the middle of the arena.
Robots are tasked to collect (virtual) items from the nest and deposit them at the source.
Retrieval and deposition of the items happen automatically, as the robots enter the respective zones.
Every successfully retrieved or deposited item is rewarded with a score of \num{10}.
Demonstrations were provided through beacon control.
Two beacons were placed in the gray area.
When entering these beacons, the robot would switch to the \textit{Come} behavior with the target within the nest and source, respectively.
The nest and source also contain beacons that, when entered, would switch the robots to the \textit{Come} behavior, with the target being the way point to the respective next goal.
Initially, the robots are selected and assigned a \textit{Come} behavior towards the closest region of interest (way point, nest, or source).

\change{We are aware, that these are basic missions and the desired collective behavior can be achieved using approaches other than imitation.
In fact, implementations of these behaviors exist readily in the literature~\cite{KaiBegPla-etal2022icra}.
We choose to focus on these basic missions to reduce the risk of unsuccessful demonstrations. This allows us to interpret the performance as the result of our proposed framework, rather than of the quality of the acquired demonstrations.}

\subsection{Real-world environment}

In addition to the simulated missions, we also consider experiments in a real-world environment.
The robot swarm operates in a $\qty{2}{\meter}\times\qty{4}{\meter}$ arena, which is enclosed by walls.
Each robot executes its own decentralized policy.
In order to prevent damage to the robots and the environment, we implemented a hardware protection layer.
Once a (potential) collision is detected, the hardware protection layer will inhibit the output of the policy and rotate in place until the obstacle is cleared.
Furthermore, the robots are tracked through a Vicon motion capture system.
This allows us to track the positions and orientations of the robots inside the arena and to compute the rewards obtained in these experiments.

\subsection{Experimental Protocol}

For each mission, we conduct five independent experimental runs.
In each run, we provide one human-operated demonstration that is used to generate five different rollouts, resulting in five trajectories of feature-action pairs.
\change{We provide demonstrations following the general strategies described above, allowing for refinement of the exact demonstrations until we were satisfied with the swarm performance.}
Each demonstration takes 51 seconds.
In our experiments, this proved to be a good tradeoff between time for the desired behavior to arise and total time necessary to evaluate all policies.
The set of five demonstration rollouts was then provided as the expert demonstrations to the imitation learning framework.
In order to reduce the time necessary to evaluate the policies, we reduced the control frequency of the robots in simulation to \qty{1}{\hertz}.
To track the development of the imitation learning process, we log information related to the discriminator, as well as a copy of the current policy in regular intervals (for a total of 100 logged policies per run).
To assess the quality of the imitated policy, it is then evaluated five times in simulation.
Additionally, we transfer it to the real robots and evaluate it once in the real world.
The control frequency for the real robots is set to \qty{10}{Hz}.

Additionally, we also consider imitation learning from trained policies.
To that end, we perform a standard reinforcement learning process for each mission using PPO and the policy architecture described in Sec.~\ref{sec:swarm-gail}.
The generated policy is then evaluated five times in simulation to generate five policy rollouts (analogous to the five human-operated demonstrations).
These rollouts are then used as inputs into an imitation learning process under the same experimental protocol as for human-operated demonstrations.
In particular, we also repeat this process five times per mission, leading to five different learned policies, one for each repetition.

All code, demonstrations, policies, performance data and videos are available online\footnote{\url{https://github.com/CPS-Konstanz/SwarmGAIL}}.

\section{Results}

\begin{figure}
    \centering
    \includegraphics[width=\linewidth]{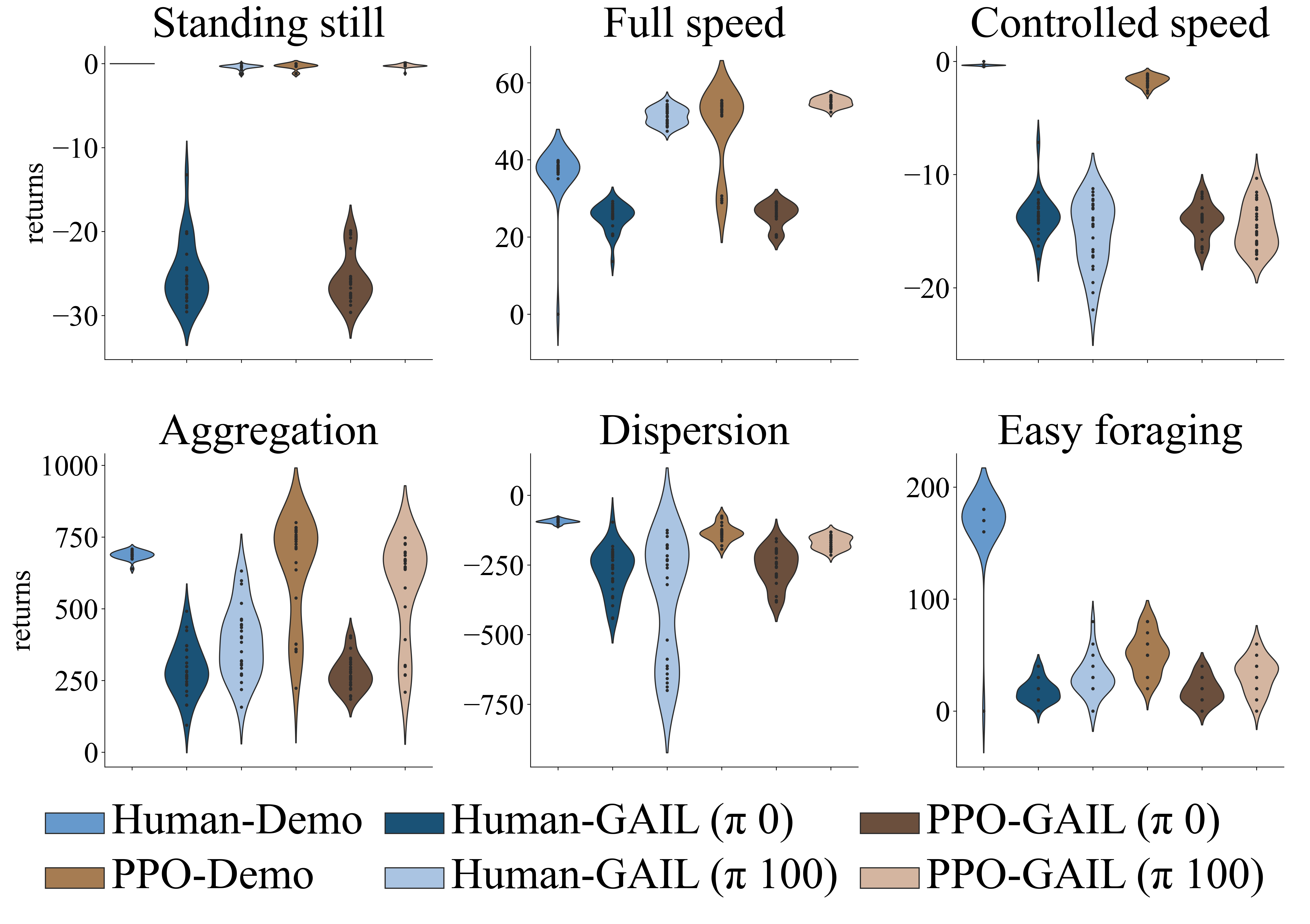}
    \caption{Violin plots of the return (cumulative reward) of all evaluations across all considered missions.
    Colors indicate the source of the demonstrations (blue from human-operated,  brown from PPO-trained ones).
    Dark colors represent the initial policies of each imitation learning experiment, light colors the final policy.
    Black circles correspond to one evaluation of one policy/demonstration.
    The higher the return, the better the performance.}
    \label{fig:quality}
\end{figure}

\subsection{Manual demonstrations vs. trained policies}
Fig.~\ref{fig:quality} shows the performance of the rollouts of all demonstrations.
As a general trend, using manual control or a trained policy results in similarly good performing behaviors.
This is in so far noteworthy, as the human demonstrations do not explicitly aim to maximize the reward function.
Differences in the quality are often results from particularities of the mission and demonstrations.
For example, in \MStandingStill, the manual demonstration easily reaches a score of exactly \num{0}, as it uses the \emph{Stop} behavior on all robots from the beginning.
The PPO-trained policy, however, has troubles setting the linear velocity to exactly \qty{0.0}{\meter\per\second} all the time.
As a result, some of the robots move from time to time, accumuluting small penalties.

In the mission \MFullSpeed, contrarily, the PPO trained policies usually discover strategies that are not available to the human demonstrator.
Our demonstrations make use of the \emph{Random walk} behavior, which has the robots performing a ballistic motion behavior.
The robots move straight with their maximal velocity until they encounter a wall, at which point they turn in place.
The turning in place results in short periods of time where the linear velocity of this robot is \qty{0.0}{\meter\per\second}, therefore not contributing to the reward.
In the PPO-trained policies, however, the robots usually drive in a circular motion.
This has the effect of keeping the linear velocity (nearly) constant, while the angular velocity is set very high to avoid collisions.
Of special note is also the mission \MForaging.
The manually provided demonstrations usually achieve scores of more than \num{150}. 
However, no rollout of the PPO-trained policy results in a score of at least \num{100}.
This observation supports our belief that a greater focus should be put on human demonstrations.
Especially in more complex missions (of which \MForaging is still a simple representative), human demonstrations can easily eclipse those behaviors generated by other methods.

Lastly, it should be noted that the human demonstrations showed less variance in their performance, compared to the PPO-trained policies.
This happens for two reasons:
First, it can happen that some of the five PPO-trained policies fail to adequately address the mission.
This was, however, never a problem with the manual demonstrations.
Second, the performance of the same PPO-policy varies more across repeated rollouts than the performance of the human operator across repeated demonstrations.
Only in the mission \MForaging, do the human-operated demonstrations show greater variance than the ones generated from a PPO-trained policy.

\subsection{Generative adversarial imitation learning}


Fig.~\ref{fig:quality} shows also the performance of the generative adversarial imitation learning process.
Besides the performance of the demonstrations (human and PPO-trained), we also show the performance of the initial policy ($\pi$ 0) and the final generated one ($\pi$ 100).
As initial policies are randomly generated, their performance only serves as a baseline to highlight the degree to which the performance improved.

For the missions \MStandingStill\ and \MFullSpeed, we observe that initial policies are, as expected, relative poor.
The final trained policies, however, match the quality of the demonstrations, and in the case of \MFullSpeed even exceed the manual demonstration.
The learned policies make use of the fact that they can control both the linear and angular velocity, while manual demonstrations relied on ballistic motion.
As previously mentioned, the ballistic motion accrues small penalties for turning in place after collisions.
In the imitating policies, robots however set their angular velocity to avoid collisions.
This strategy is similar to the one used in PPO-trained policies.
Yet, the policies trained from human-operated demonstrations, turn less than the PPO-trained policies, which essentially turn in place.
This is due to the fact that the ballistic motion covered the full arena, affecting swarm-level features such as \emph{grouping} or \emph{coverage}.
Therefore, also the imitating policies cover the complete arena.
Accordingly, the policies imitating the PPO-trained demonstrations learned also to turn nearly on the spot, mimicking closely their demonstrations.
This shows that the imitation learning process was able to pick up on these idiosyncrasies in the demonstrated behavior, which are reproduced in the corresponding imitating policies.
\change{Yet, the fact that GAIL was not able to correctly imitate the turning-in place of ballistic motion warrants further investigation, even if on paper it resulted in a better performing behavior.}

In the case of \MControlledSpeed, the imitating policies do not reach the same performance as the demonstrations.
In fact, when compared to the initial policies, the imitation process seems to have affected the performance negatively.
Visual inspection of the initial and final policies indicate that both appear to rely on controlling their linear velocity through randomly stopping and moving.
This incurs high penalties, as the average velocity is then most of the time either too low or too high.
The human-operated and PPO-trained demonstrations, however, move constantly at nearly the target velocity.
It is not clear, why the demonstrated behavior did not emerge in either imitation learning process.

In \MAggregation, the GAIL runs, that used the rollouts from PPO-trained policies as demonstrations, were able to learn to imitate the behavior well.
The policies imitating human demonstrations do improve over the random initial ones but do not score as well as the originals.
Visual inspection of the generated behaviors reveals that robots are mostly stationary and only try to slowly approach each other.
The score of the behavior is therefore highly contingent on the initial starting positions of the robots.
This might be caused by the interplay of swarm-level features and manual demonstrations.
The human operator provides a quick aggregation, after which the robots remain stationary.
This results in most of the features to develop similarly to a stationary swarm (except for the early increased \emph{average speed}).
It seems difficult for the imitation learning process to copy this aggregation but it seems to focus mostly on the longer, stationary part of the demonstration.
In the PPO-trained demonstrations and the resulting imitating policies, however, the aggregation is not static but dynamic (except for one pair of policies that also learn the static aggregation).
This generates a more dynamic representation in the swarm-level features, which apparently is more helpful to the imitation learning process.


In the mission \MDispersion, the imitation learning improves only slightly over the random initial policies.
They, however, show less variance in their performance (with the exception of one policy that appears to have learned an aggregation behavior).
This can be explained by the fact that the initial policies do not contain very structured motion.
As a result, their score is mostly dependent on the initial position.
The learned policies provide more structured behavior that allows the robots to separate, should they have started too close together.
Yet, as the LiDAR readings of the robots are limited to \qty{2}{\meter}, the robots can only separate themselves until they don't perceive each other.
The human demonstrator could go one step further and position the robots maximally spread out in the arena.

In the last considered mission, \MForaging, the human demonstrations clearly outperform the imitating policies.
Our initial belief was, that the robots might learn to move in circles/spirals that overlap with the source and the nest (and the manual demonstrations draw inspiration from this strategy as well with their way points).
However, the initial policies usually generate a slow, randomly moving behavior.
The final trained policies (both from human-operated demonstrations and PPO-trained ones) exhibit a much faster, but nonetheless randomly moving behavior.
The increased speed of movement makes it more likely to retrieve and store items, but those occurrences remain accidental.
As also the PPO-trained demonstrations fail to exhibit any behavior that appears to be meaningfully different from a random walk, we believe this might be an indication that the environment lacks structure that the robots can exploit for navigation.
Additionally, our swarm-level features related to the colors might have further complicated the imitation learning process.
The manual demonstrations easily manage to navigate between floor patches, resulting high \emph{visit frequencies} and low \emph{color travel times}.
As navigation for the learned policies is much harder, they are unable to match these features.
As a result, the discriminator can very easily distinguish between demonstrated and learned behaviors.
Therefore, the reward landscape for the policy improvement is flat, as no possible improvement will overcome the discriminator.

\subsection{Real-world validation}

\begin{figure}
    \centering
    \includegraphics[width=\linewidth]{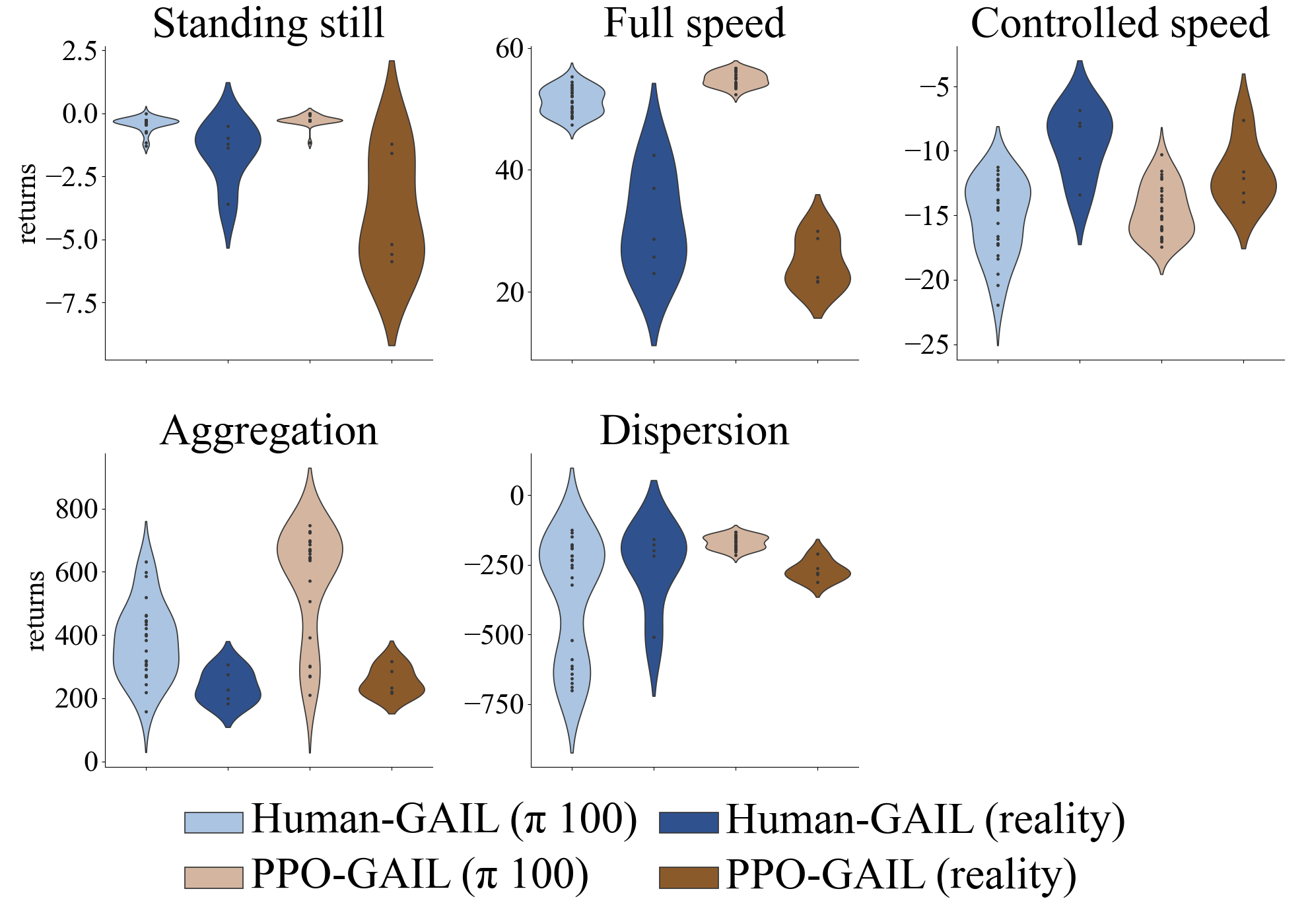}
    \caption{Violin plots of the returns (cumulative reward) of imitated policies in simulation (light colored) and reality (dark colored).
    Colors indicate the source of the demonstrations (blue from human-operated ,  brown from PPO-trained ones).
    Black circles correspond to one evaluation of one policy/demonstration.
    The higher the return the better the performance.}
    \label{fig:quality-realworld}
\end{figure}

Fig.~\ref{fig:quality-realworld} shows the performance of the experiments with real robots.
We excluded the mission \MForaging from consideration for two reasons.
First, the real TurtleBot~4 robots did not have a ground sensor that would have been necessary to perform this mission.
While we could have provided a virtual ground sensor, we also changed the arena size.
The policies were not particularly well-performing in simulation and the added change of arena size would probably have impacted the performance even further.

It can be seen that the policies generated from manual and PPO-trained demonstrations perform similarly, when evaluated on the real robots.
This is not surprising, as these policies already performed relatively similar in simulation.
Notably, visual inspection of the robot swarm reveals that each mission results in clearly distinguishable and recognizable behavior (with the exception of the one dispersion policy that resulted in an aggregation-like behavior).
This further proves that our framework was able to successfully imitate the demonstrated behaviors and the generated behaviors remain qualitatively robust to the reality gap.

Quantitatively, the policies are affected in different ways by the reality gap.
The biggest impact on behavior is generated by the hardware protection layer.
In some missions, such as \MAggregation and \MFullSpeed, the hardware protection layer prevents collisions that are a considered an acceptable part of the behavior in simulation.
As a result robots slow down and turn away or stop farther from each other than would happen in simulation.
Interestingly, this same effect seems to have been beneficial for the mission \MControlledSpeed, where robots performed better in reality than in simulation.
The performance in \MStandingStill and \MDispersion was not significantly affected by the hardware protection layer, as robots did not encounter each other often during the experimental runs.
This leads us to believe that our method might transfer well to real robots, if the hardware protection layer would have been appropriately modelled in the simulation.

\section{Conclusion}

In this work, we have presented an imitation learning framework for robot swarms.
We model the imitation learning problem as a single-agent problem and solve it using GAIL.
The discriminator classifies samples according to five swarm-level features and the policy controls robots according to their own local observations.
We have also provided a tool that allows us to provide human-operated demonstrations.
Our results show that for all missions except \MControlledSpeed and \MForaging, the imitation learning process was able to learn qualitatively meaningful behaviors that performed similarly well as the provided demonstrations.
In addition, we have tested the policies on real robots.
The transfer of the policies went relatively well with the behaviors being still visually recognizable on the real robots.
However, the hardware protection layer on the real robots was not modeled in simulation and therefore impacted the performance of the transferred policies.

Our work has shown, however, also some limitations of our approach.
\change{Despite considering relatively simple missions, i}n some cases, 
the demonstrated behavior could not always be accurately imitated.
We believe that this is partially due to our ad-hoc selected swarm-level features, which impacted the imitation learning process\change{, however, more research on identifying the limiting factors is needed}. 
\change{Another potential factor is the increased performance variance by both human and PPO-trained demonstrations in the more complex missions, such as \MForaging.
We believe that more work is needed in order to provide reliable demonstrations for these complex missions.}
\change{In addition, f}uture work should be devoted to better understanding how the features impact the imitation learning process \change{(e.g., by performing ablation studies)} and how researchers can systematically select these features to ensure proper performance of the imitated behaviors.
\change{The fact that we used the same features as input for the discriminator and the reward function is another decision that warrants further consideration.
This could lead to instances of information leaking, in which the discriminator implicitly learns which features were associated with the rewards function assumed to be maximized by the demonstration.
In future work, a more decoupled approach to feature-space and reward function should be considered.
This applies in particular, when moving to more complex missions.}

\change{Despite these limitations, our approach also opens up new avenues for research.
We believe that it would be of interest to compare this approach against other methods from the literature, to source demonstrations from experts with varying levels of expertise in swarm robotics, and to study how well the trained policy scales to different swarm sizes.}


\section*{ACKNOWLEDGMENT}
%
We thank Mahabaleshwar Ammu 
for their help in setting up the experiments with real robots.


\bibliographystyle{IEEEtran}
\bibliography{demiurge-bib/definitions,demiurge-bib/address,demiurge-bib/author,demiurge-bib/institution-short,demiurge-bib/journal-short,demiurge-bib/proceedings-short,demiurge-bib/publisher,demiurge-bib/series-short,demiurge-bib/bibliography,additions}

%

\end{document}